%% file: main.tex
\documentclass[letterpaper, 10 pt, conference]{ieeeconf} 
\IEEEoverridecommandlockouts                             
\usepackage[backend=biber,
            hyperref=true,
            url=false,
            isbn=false,
            doi=false,
            backref=false,
            style=ieee,
            citestyle=numeric-comp,
            sorting=nyt,
            block=none]{biblatex}

\usepackage{flushend}
\usepackage{balance}

\usepackage{amsmath}
\usepackage{float}
\usepackage{booktabs}
\usepackage{graphicx}
\usepackage{hyperref}
\usepackage{multirow}
\usepackage{balance}
\usepackage{placeins}
\usepackage{soul}
\usepackage{standalone}
\usepackage[font={small}]{caption}
\usepackage{subcaption}
\usepackage{tikz}
\usepackage{url}
\usepackage{xcolor}
\usepackage[utf8]{inputenc}

\title{\LARGE \bf
Hierarchical Planning for Long-Horizon Manipulation\\with Geometric and Symbolic Scene Graphs
}

\author{Yifeng Zhu$^{1,2}$, Jonathan Tremblay$^{1}$, Stan Birchfield$^{1}$, Yuke Zhu$^{1,2}$
\thanks{$^{1}$NVIDIA, $^{2}$The University of Texas at Austin.}
}

\newcommand{\pick}{\texttt{pick}}
\newcommand{\place}{\texttt{place}}
\newcommand{\push}{\texttt{push}}

\newcommand{\InGrasp}{\texttt{InGrasp}}
\newcommand{\Clear}{\texttt{Clear}}
\newcommand{\On}{\texttt{On}}
\newcommand{\In}{\texttt{In}}
\newcommand{\fpreimage}{$f_{\text{preimage}}$}
\newcommand{\freach}{$f_{\text{reach}}$}

\DeclareUnicodeCharacter{0301}{\'{e}}

\begin{document}

\maketitle
\thispagestyle{empty}
\pagestyle{empty}

\begin{abstract}
We present a visually grounded hierarchical planning algorithm for long-horizon manipulation tasks. Our algorithm offers a joint framework of neuro-symbolic task planning and low-level motion generation conditioned on the specified goal. At the core of our approach is a two-level scene graph representation, namely \emph{geometric scene graph} and \emph{symbolic scene graph}. This hierarchical representation serves as a structured, object-centric abstraction of manipulation scenes. Our model uses graph neural networks to process these scene graphs for predicting high-level task plans and low-level motions. We demonstrate that our method scales to long-horizon tasks and generalizes well to novel task goals. We validate our method in a kitchen storage task in both physical simulation and the real world. Experiments show that our method achieves over 70$\%$ success rate and nearly 90$\%$ of subgoal completion rate on the real robot while being four orders of magnitude faster in computation time compared to standard search-based task-and-motion planner.\footnote{Additional material at \url{https://zhuyifengzju.github.io/projects/hierarchical-scene-graph}.}

\end{abstract}

\section{Introduction}
\label{sec:intro}
\input{1-intro}

\section{Related Work}
\label{sec:related_works}
\input{2-related_works}

\section{Problem Statement}
\label{sec:problem}
\input{3-problem}

\section{Method}
\label{sec:methods}
\input{4-methods}

\section{Experiments}
\label{sec:experiments}
\input{5-experiments}

\section{Conclusions}
\label{sec:conclusions}
\input{6-conclusion}

\section*{Acknowledgement}
We would like to thank Guanya Shi for providing an improved version of
DOPE. We also would like to thank Nvidia AI-Algorithm team for providing
valuable internal feedback to the paper.

\printbibliography
\clearpage

\end{document}

%% file: 1-intro.tex




Long-term robot autonomy in everyday environments demands a robot to make autonomous decisions over prolonged time periods. It is typically cast as a sequential decision making problem where a robot searches for feasible \emph{plan}s as sequences of actions in a finite horizon. In robot manipulation, a fundamental challenge stems from the high-dimensional search space of continuous robot actions, exacerbated by the combinatorial growth of planning complexity with respect to the task length.

A common approach to tackling long-horizon tasks is task-and-motion planning (TAMP) that factorizes the planning process into discrete symbolic reasoning and continuous motion generation~\cite{kaelbling2011hierarchical}. However, several challenges present in existing methods hinder their applicability in real-world domains: 1) these methods rely on manually specified symbolic rules, namely planning domains; 2) they typically abstract away raw perception with known physical states; 3) searching with discrete symbols and continuous motions remains prohibitively expensive in complex domains.

Over the past few years, a rising number of data-driven methods have been developed for robotic planning. Most notably, deep learning techniques have demonstrated their effectiveness in end-to-end learning of sensor-space models~\cite{Finn2017DeepVF,oh2015action} or latent-space models~\cite{fang2019cavin,hafner2019dream,watter2015embed} for planning and control. These methods excel at integrating raw sensory input and generating reactive behaviors, but fall short of the generalization capabilities for long-horizon tasks in comparison to TAMP algorithms. This dichotomy has ignited a series of investigations into integrating learning into the TAMP paradigm to speed up the search of feasible plans~\cite{chitnis2016guided,kim2019learning,wang2018active} or to directly predict the plan feasibility from an initial image~\cite{driess2020deep}. However, these models still rely on strong domain knowledge and costly search procedures required by TAMP methods.

\begin{figure}[t]
    \centering
    \includegraphics[width=1.0\linewidth]{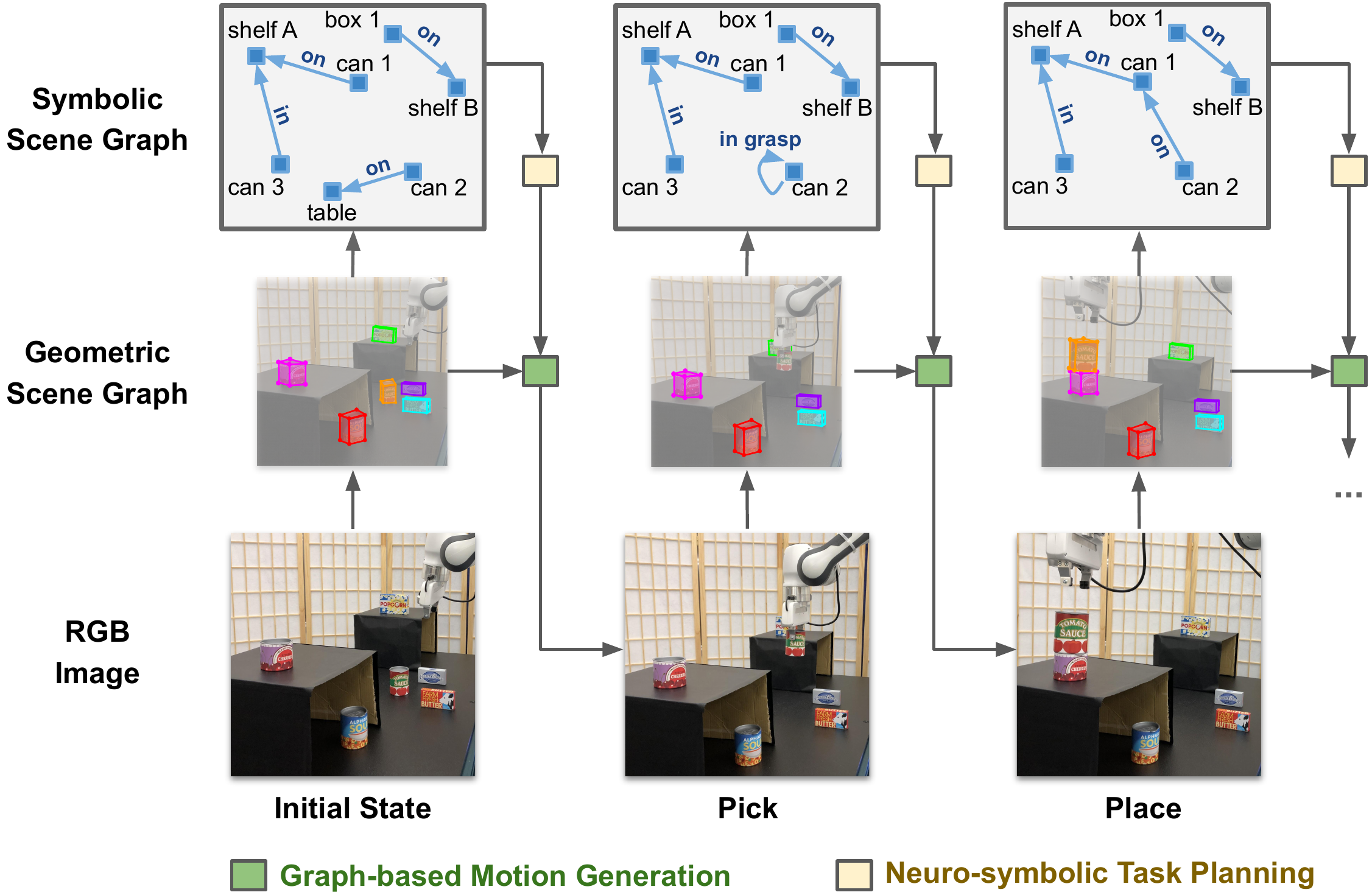}
    \caption{We introduce a hierarchical planning model for vision-based long-horizon manipulation. We represent a manipulation scene with object-centric scene graphs at both symbolic and geometric levels, which can be used for neuro-symbolic task planning and graph-based motion generation, respectively.} 
    \label{fig:pull_figure}
    \vspace{-5mm}
\end{figure}

To tackle the challenges present in the long-horizon planning problem, we introduce a visually grounded hierarchical planning algorithm for long-horizon manipulation tasks. Our algorithm operates directly on the visual observations and performs high-level task planning and low-level motion generation conditioned on a specification of the task goal. Through a novel combination of neural network-based learning and symbolic reasoning, this approach eliminates the need for predefined symbolic rules and complex motion planning procedures in TAMP. And by operating on our proposed graph-based representations, it attains a form of strong generalization to longer task instances which are difficult for end-to-end learning approaches.

At the heart of our approach is the structured object-centric representation of visual scenes. We propose to represent a manipulation scene as a pair of \emph{scene graph}s~\cite{armeni20193d,rosinol20203d}, a two-level abstraction shown in Figure~\ref{fig:pull_figure}. The low level is a \emph{geometric} scene graph depicting the 6-DoF poses of the entities in the environment and their relative spatial relations. The high level is a \emph{symbolic} scene graph that describes the abstract semantic relations among the entities and the robot. We obtain the two-level object-centric scene graphs of real-world household objects with a state-of-the-art 6-DoF pose estimation method from RGB cameras~\cite{tremblay2018corl:dope} and symbol mapping. This hierarchical scene graph representation operates along with our hierarchical planning algorithm. 

At each step in the manipulation process, the current symbolic scene graph is used for a neuro-symbolic regression planner to determine the next immediate subgoal from the final goal specification. The regression approach frees us from massive search time while not requiring predefined symbolic rules as in TAMP. Then the subgoal is used for grounding with geometric scene graphs, opening the door for predicting motion parameters to reach the subgoal directly on this grounded geometric scene graph using graph-based motion generation. To enable neuro-symbolic task planning and graph-based motion generation, we integrate recent advances in learning-based neuro-symbolic reasoning~\cite{cranmer2020discovering,danfei2019regression} and object-centric dynamics model learning~\cite{kipf2019contrastive,ye2019ocmpc}. Both the task planning and motion generation models are based on Graph Neural Networks (GNNs)~\cite{hamrick2018relational} which are deep learning modules for graph processing. 



\textbf{Summary of Contributions:} 1) We introduce a visually-grounded hierarchical planning algorithm that scales to long-horizon tasks through the use of two-level scene graph representations; 2) We design graph-based learning models and training procedures that enable our models to generalize to novel tasks; and 3) We demonstrate the effectiveness of our approach in tabletop manipulation tasks in both a simulated environment and the real world.

%% file: 2-related_works.tex
\noindent
\textbf{Task and Motion Planning.}
Our approach to factorizing long-term manipulation planning into the task level and the motion level has a strong resemblance to TAMP methods~\cite{dantam2016incremental,kaelbling2011hierarchical,migimatsu2020object,Srivastava2014CombinedTA}.  Classic TAMP approaches rely on predefined symbolic rules and known dynamic models. They perform discrete task plans with logic search and continuous motion plans through sampling-based techniques~\cite{garrett2020pddlstream,kaelbling2011hierarchical} or trajectory optimization~\cite{migimatsu2020object,toussaint2018differentiable,toussaint2015logic}. TAMP methods can generalize to arbitrary goals defined in a problem domain; however, they suffer from the combinatorial complexity of searching for the feasible plans. Learning techniques~\cite{chitnis2016guided,kim2019learning,wang2018active} have been increasingly employed into the TAMP framework to accelerate inference in replacement of the handcrafted search heuristics. They aim to guide the search procedure toward more promising task plans, reducing the number of motion planning problems required to be solved. Alternatively, recent work has explored learning techniques to evaluate the feasibility of plans from visual observations~\cite{driess2020deep,wells2019learning}. Most relevant to us is Deep Visual Reasoning~\cite{driess2020deep} which aims to directly predict task plans from initial scene images. However, it predicts the task plan solely based on the initial image, and relies on environment dynamics for costly trajectory optimization. In contrast, our model performs regression task planning and achieves high inference efficiency by predicting both the task plans and motion plans with graph neural network-based models.

\vspace{1mm}
\noindent
\textbf{Learning to Plan.}
This work is also closely related to recent efforts on learning to plan with end-to-end deep learning~\cite{fang2019cavin,Finn2017DeepVF,hafner2019dream,hafner2019learning}. The core idea is to learn predictive models from data which can be used for model-based reinforcement learning or planning. The dynamic models can be learned in raw sensor space~\cite{Finn2017DeepVF,nair2019hierarchical,oh2015action} or learned latent space~\cite{fang2019cavin,hafner2019dream,hafner2019learning}. A major advantage of these learning models is that they can integrate seamlessly with neural representations learned from real-world sensory data and require less domain knowledge than TAMP methods. However, the ability of accurate long-term predictions requires a strong generalization capability generally beyond generic neural networks. To mitigate this problem, hierarchical structures have been introduced to neural network methods~\cite{fang2019cavin,lynch2020learning,nair2019hierarchical,nasiriany2019planning,pertsch2020long}. Like ours, these methods decompose a plan into high-level subgoals and low-level motions. Harnessing the temporal abstraction, these methods are able to make long-horizon predictions for multi-stage tasks and generalize well to tasks within the training distributions. However, the neural representations suffer from their limited generalization power to out-of-distribution or longer tasks. Inspired by the TAMP structure, recent learning work has looked into neuro-symbolic hybrid models for robotic planning~\cite{huang2019continuous,huang2020motion,danfei2019regression} that exploits the complementary strengths of neural representations and symbolic representations. Among these methods, Regression Planning Network (RPN)~\cite{danfei2019regression} extends classical regression planning in symbolic space with deep networks, which achieves strong zero-shot generalization to unseen tasks. However, RPN simplifies perception as 2D object bounding boxes and abstracts away the low-level motions. Our model adopts the similar regression planning on the task level, and moreover, directly interfaces with a low-level motion generation module through the hierarchical scene graphs from real images.

\vspace{1mm}
\noindent
\textbf{Graph-based Scene Representations.}
A key factor determining the generalization capability of a planning model is the choice of state representations. Many works have demonstrated that factorized representations of objects and entities improve a model's ability to reason about novel scenes and new tasks~\cite{burgess2019monet,engelcke2019genesis,lin2020space}. For robot manipulation, the entities and their relations in the environment supply essential information for planning a robot's actions. This has motivated a variety of graph-based scene representations~\cite{armeni20193d,krishna2017visual,rosinol20203d,sieb2020graph,xu2017scene} developed in the robotics and computer vision community. These representations have been used in prior work of robotic manipulation for short-term skill learning~\cite{sieb2020graph,ye2019ocmpc} or high-level task planning~\cite{danfei2019regression}. Instead, we integrate the scene graph representations at the two levels of abstraction to offer a coherent framework of task planning and motion generation. In a similar vein, different forms of hierarchical 3D scene graph representations~\cite{armeni20193d,rosinol20203d} have been developed in the 3D vision and SLAM communities. Our key novelty, in comparison, is to make use of hierarchical representations for visual planning.

%% file: 3-problem.tex
We consider a vision-based goal-conditioned planning problem. Let $\mathcal{O}$ be the space of raw visual observations, $\mathcal{A}$ be the action space of robot motor control, and $\mathcal{G}$ be the space of goal specifications, the goal of our algorithm is to find a policy $\pi: \mathcal{O} \times \mathcal{G} \rightarrow \mathcal{A}$ to generate low-level motor action $a \in \mathcal{A}$ given the current observation $o \in \mathcal{O}$ and a goal $g\in \mathcal{G}$. In complex manipulation tasks, it takes a prolonged sequence of low-level actions to achieve a long-term goal, making the planning problem computationally prohibitive. To overcome this challenge, we factorize the problem into high-level task planning and low-level motion generation. 

We design two-level scene graphs as the representation for our factorization of the problem (see Figure~\ref{fig:overview}). On the low level, a \emph{geometric scene graph} $s_{g}$, similar to kinematic graph~\cite{lavalle2006planning}, describes the Cartesian locations of objects. Each node defines the pose of an object in \emph{SE}(3), and each edge defines the relative spatial relations between two objects. A geometric scene graph $s_g$ can be computed from a visual observation $o\in\mathcal{O}$ with a 6-DoF \emph{pose estimation} model $f_p$ where $s_g=f_p(o)$. On the high level, we use a vocabulary of semantic symbols to form an abstract description of a scene, allowing us to express both the current state and the goal specification with logical predicates. A \emph{symbolic scene graph} $s_s$ is thus a set of visually grounded predicates that characterizes the object states and the semantic relations between objects. We refer to nodes or objects interchangeably. 

Given a geometric scene graph, we can design a \emph{symbol mapping} function $f_m$ to compute its corresponding symbolic scene graph $s_s=f_m(s_g)$. Let $\mathcal{Z}$ be the space of all possible sets of predicates. In this case, we have $s_s\in\mathcal{Z}$ and $\mathcal{G}\subset\mathcal{Z}$. Our hierarchical planning process is to learn a task planning model $\phi$ and a motion generation model $\psi$ where
\begin{align}
    \phi &: \mathcal{Z} \times \mathcal{G} \rightarrow \mathcal{G} \label{eq:regresion-planning}\\
    \psi &: \mathcal{O} \times \mathcal{G} \rightarrow \mathcal{A} \label{eq:motion-geneartion}.
\end{align}
The task planning model $\phi$ takes a symbolic scene graph $s_s\in\mathcal{Z}$ and a task goal $g\in\mathcal{G}$ as input, and predicts the next subgoal $\hat{g}\in\mathcal{G}$, namely an immediate subgoal that can be achieved by a single motion primitive (see Section~\ref{sec:task-planning}). Subgoals are the necessary intermediate states to be achieved before reaching the goal $g$. The motion generation model $\psi$ computes $s_g$ from $o\in\mathcal{O}$, processes $s_g$ and $\hat{g}$ with a \emph{symbol grounding} function $f_g$, and generates a low-level action $a\in\mathcal{A}$ (see Section~\ref{sec:motion-generation}).

%% file: 4-methods.tex
\begin{figure*}[ht]
    \centering
    \includegraphics[width=0.8\linewidth]{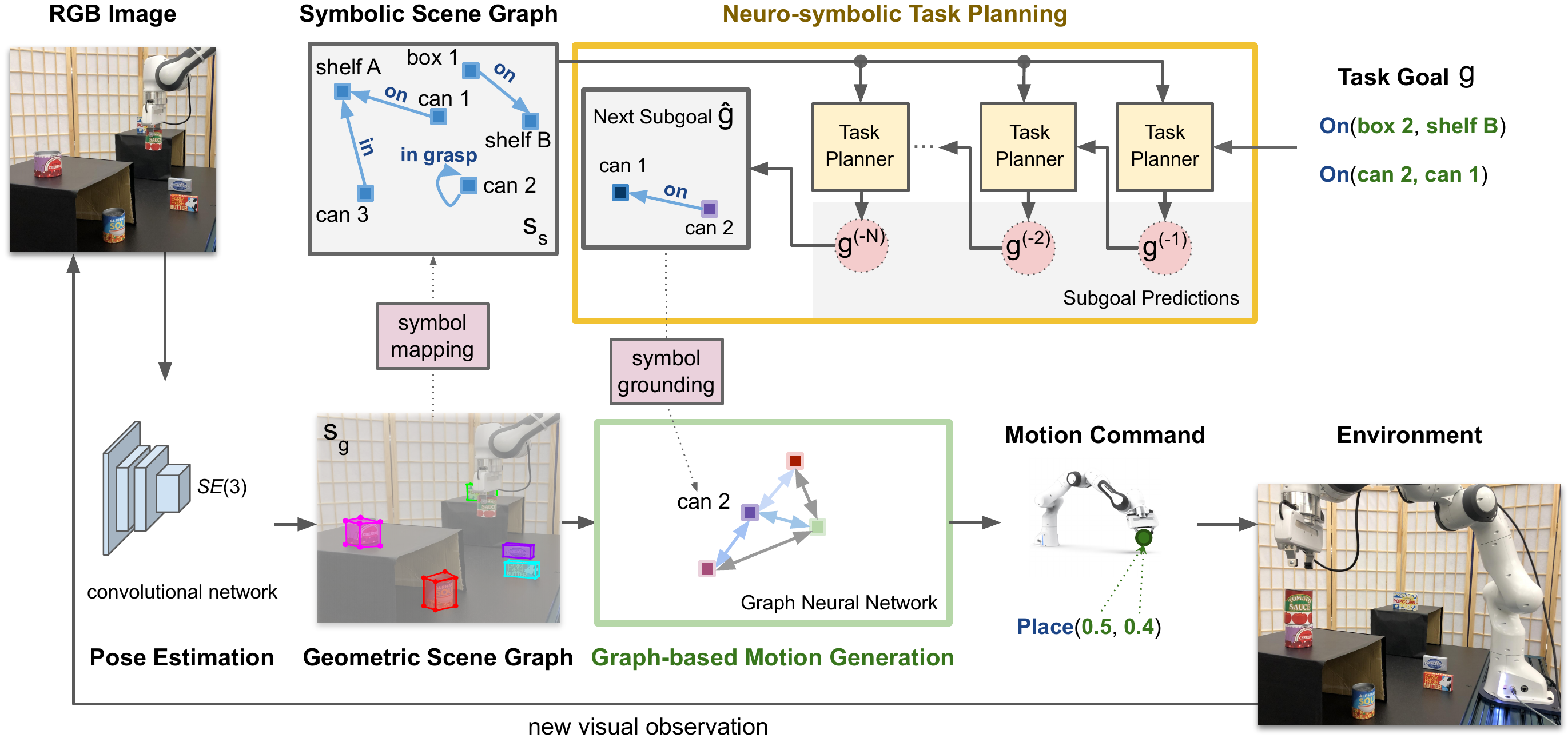}
    \vspace{-1mm}
    \caption{An overview of our scene graph-based  hierarchical planner. 1) Given an observation $o$ of the environment, we use a CNN based pose estimator $f_p(o)$ to obtain poses of objects. 2) We construct the geometric scene graph $s_g$ using output from pose estimation. Edge connections are omitted from the graph in the figure. 3) A symbol mapping function $f_m$ maps from $s_g$ to $s_s$. 4) A symbolic goal is given to the task planner, and the task planner recursively predicts subgoals $g^{-i}$ until the next immediate subgoal $\hat{g}$. 5) The geometric scene graph $s_g$ is grounded with next immediate subgoal $\hat{g}$ and motion generation model takes in the grounded graph and the next immediate subgoal to command low-level motor control for robot to execute a primitive such as \pick, \place, or \push.}
    \label{fig:overview}
    \vspace{-4mm}
\end{figure*}
An overview of our method is shown in Figure~\ref{fig:overview}. From the current RGB observation, our model constructs a hierarchical representation of the geometric scene graph and the symbolic scene graph. At the high level, it performs neuro-symbolic task planning and predicts the next symbolic subgoal conditioned on the current symbolic scene graph and the specified goal. At the low level, it performs forward motion generation by grounding the symbolic subgoal to the current geometric scene graph and predicting motion parameters. Once the next subgoal is achieved, it replans according to this procedure with the new observation. As both levels directly operate on structured scene graphs, it makes Graph Neural Networks (GNNs)~\cite{hamrick2018relational} a natural choice for our model architecture. GNNs are a family of geometric deep learning models for graph processing with the relational inductive biases for stronger generalization. A GNN learns features for each graph node and for each graph edge with parameter-shared modules (neural networks) and uses a message passing mechanism to exchange information among nodes in the local neighborhood. This design enables the model to aggregate contextual information through the graph structure while generalizing to different graph topologies, such as different number of objects in our scene graphs.

\subsection{Task Planning}
\label{sec:task-planning}
The task planning model $\phi$ is built on top of Regression Planning Networks (RPN)~\cite{danfei2019regression}, which performs neuro-symbolic regression (backward) planning in the symbolic space using deep networks. Given a specification of the final task goal $g$, we denote its previous subgoal $g^{(-1)}$, and similarly, $g^{(-i-1)}$ as the previous subgoal of a subgoal $g^{(-i)}$. The core idea of regression planning is to recursively predict $g^{(-1)}$, $g^{(-2)}$ from $g$ and $s_s$, all the way till a predicted subgoal $g^{(-N)}$ is \emph{reachable}, that is, this subgoal can be achieved from the current state with a single motion primitive (see Section~\ref{sec:motion-generation}). We refer to this next immediate subgoal as $\hat{g}$. This approach for task planning has two major benefits: 1) Regression planning searches in a backward fashion, significantly reducing the search space compared to forward search. 2) The model can be trained on demonstrations of shorter tasks without predefined symbolic rules while generalizing to longer unseen tasks.

The task planning model $\phi$ has three key components: preimage network \fpreimage, subgoal serialization, and reachability network \freach. \fpreimage~takes a goal $g$ or a subgoal $g^{(-i)}$ as inputs, and predicts all possible previous subgoals. This is achieved by predicting every edge relation in symbolic scene graph that needs to be satisfied preceding current subgoal in query, forming a preimage symbolic scene graph. 
\fpreimage~is a GNN model which uses a message passing mechanism to predict relations between every pair of nodes in the preimage graph. The subgoal serialization function then takes in all possible subgoals from the preimage graph, and selects one subgoal for subsequent reachability check. In contrast to RPN which uses a dependency network for selection, we uniformly sample a next subgoal among all possible next subgoals as we find this effective.

 After a possible subgoal is selected, \freach~checks if this subgoal is reachable from the current state with a single motion primitive. Specifically, we check whether every target semantic relation of objects in the subgoal can be achieved. \freach~is also a GNN model, which predicts a binary label of whether the target semantic relation can be achieved. If the next subgoal is reachable, a motion primitive is selected based on the next subgoal. Otherwise, the whole process is repeated again, using $g^{(-i)}$ as inputs to \fpreimage~ until finding a reachable immediate subgoal $\hat{g}=g^{(-N)}$.

\subsection{Motion Generation}
\label{sec:motion-generation}
The motion generation model $\psi$ operates on the graph representations, including the immediate subgoal $\hat{g}$ and the current geometric scene graph $s_g$, to compute a low-level motor command $a$ to execute. We use a set of parametrized motion primitives $\pi_{\theta}$ to represent low-level motions (see details in Section~\ref{sec:experiments}). The action space $\mathcal{A}$ thus contains the continuous values $\theta$ that parametrize the motion primitives.

Our model first selects a motion primitive type and the object to manipulate based on the next symbolic subgoal $\hat{g}$. This can be done by a simple and effective heuristic that we adopt in this work. We then ground $\hat{g}$ to the geometric scene graph $s_g$ by computing new, object-centric attributes for each node relative to the object or target region in $\hat{g}$. This results in a grounded geometric scene graph $\hat{s}_g$, where the node attributes that are specific for the next subgoal $\hat{s}_g$ capture contextual information of other objects, which is important for the motion primitive to avoid collisions and determine the execution trajectories given the constraints imposed by the environment. We encode these constraints of the environment using a GNN model on $\hat{s}_g$.

The input to our motion generation model is comprised of the grounded geometric scene graph $\hat{s}_g$ and the next immediate subgoal $\hat{g}$. The model outputs parameters $\theta$ for the selected motion primitive. To predict $\theta$ conditioned on $\hat{s}_g$ and $\hat{g}$, we first compute embedding vectors for $\hat{s}_g$ using a GNN and $\hat{g}$ using a fully-connected layer nework separately, and then concatenate two embedding vectors which are passed into a multilayer perceptron (MLP) to predict $\theta$. Node attributes are specific to task design and motion primitive types. For different motion primitives, we design different object-centric node attributes in $\hat{s}_g$. 

\subsection{Scene Graph Generation}
Our scene graphs contain both predefined environment fixtures, such as table and shelves, and small dynamic objects, such as cans and boxes, that can be manipulated by the robot. The poses of the dynamic objects are computed through a 6-DoF pose estimator. The pose estimator is a convolutional neural network, which takes an RGB image as input and predicts the \emph{SE}(3) pose of an object. 
We thus construct a geometric scene graph $s_{g}$ by estimating the poses of individual objects, where each node represents an object, and its node attributes represent the object's 6-DoF pose and its object type. Geometric scene graph is a fully connected graph (all nodes are connected to each other), so that message passing in GNN can be done across all nodes, fully encoding constraints in the environment.

Once we have the geometric scene graph $s_g$, the symbol mapping function $f_m$ generates symbolic predicates for edges in $s_g$ and ignores geometric information in every node, forming the corresponding symbolic scene graph $s_s$. This symbol mapping also adds information of environment fixtures and robot's state into $s_s$ for generating semantic relations. We leverage the robot's state such as whether an object is being held by the robot in $s_s$ to compute these robot-related predicates. In practice, $f_m$ can be either manually defined or learned~\cite{huang2019continuous}. For our manipulation domains, we define the function based on heuristics for simplicity.  
In the resultant graph, each node describes the object type and each edge describes the symbolic relations. As semantic relations are usually asymmetric, we use directed edges to represent symbolic relation between object nodes and subject nodes. To represent unary predicates, we add self edges to the nodes. We provide details about the logical predicates used in our tasks in Section~\ref{sec:environment}.

\subsection{Model Training}
We train our task planning module using a set of short task demonstrations. For simplicity, we use an algorithmic planner to automate the process of generating demonstrations. Once collecting trajectories of demonstrations, we generate symbolic scene graphs of each state, obtaining the state transitions in symbolic space. For training \fpreimage, we use pairs of subsequent symbolic scene graphs from demonstration sequences. As mentioned in \fpreimage~design, it predicts the preimage graph of a subgoal. Thus during training, we directly provide previous symbolic scene graph as supervision for training. For training \freach, we use pairs of subsequent symbolic scene graphs as well, but as we only have positive examples from demonstration, we augment the data with negative examples by altering predicates of edges in the next symbolic graphs. Our label for reachability training is binary for each edge, predicting whether a symbolic relation between two nodes or the unary predicate of a node is reachable from the current symbolic state.

To train our motion generation model $\psi$ to predict parameters $\theta$ for motion primitives $\pi_{\theta}$, we follow the self-supervised training procedure similar to CAVIN~\cite{fang2019cavin}, learning from task-agnostic interactions. We procedurally generate random configurations of the environments in physical simulation, namely placements of objects, and collect training samples by sampling parameters of motion primitives and executing the sampled actions in simulation. The actions in successful execution are taken as training samples.

The pose estimator we use is DOPE~\cite{tremblay2018corl:dope}, which is trained on a synthetic dataset of photo-realistic images. Each model predicts the \emph{SE}(3) pose of a known object from a single RGB image. With domain randomization during training, it produces reliable pose estimations on real-world images.

%% file: 5-experiments.tex

\subsection{Environment}
\label{sec:environment}
\begin{figure}[t]
\centering
    \begin{minipage}[t]{0.29\linewidth}
        \includegraphics[width=\linewidth,trim=0cm 0cm 0cm 0cm,clip]{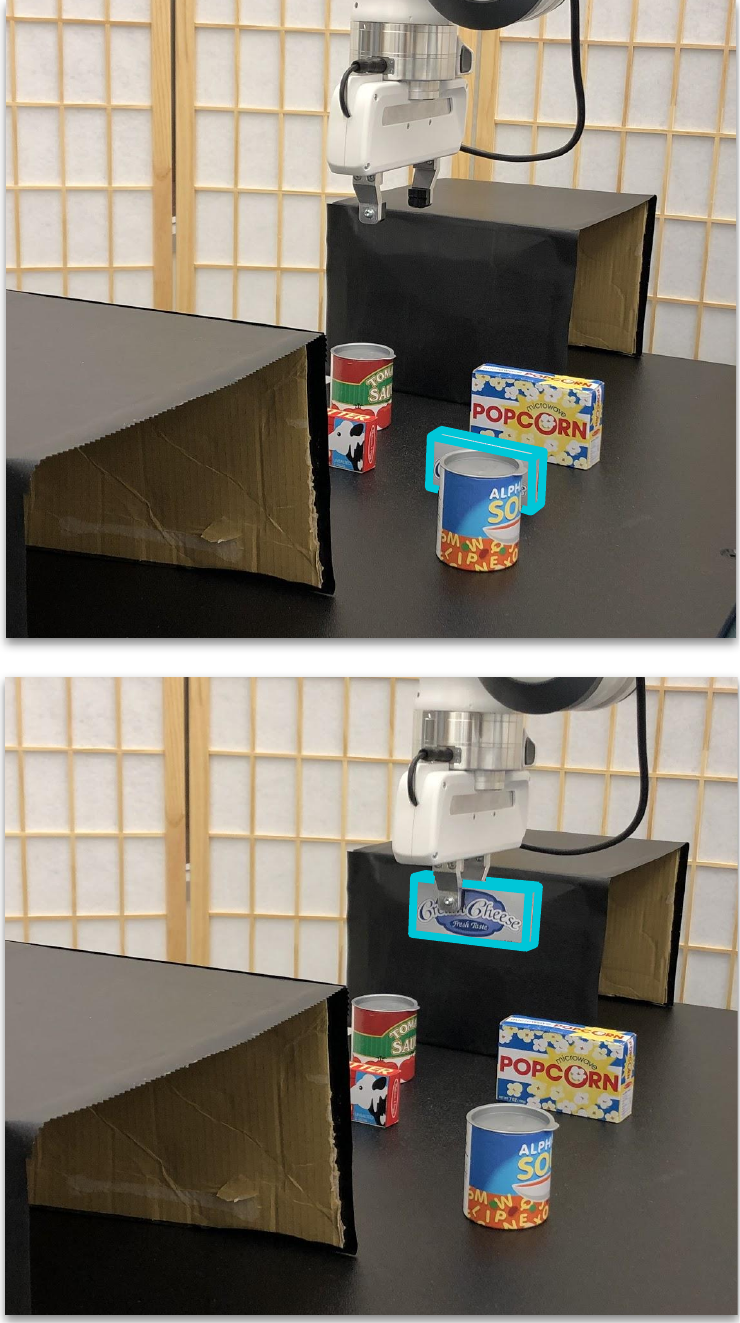}
       \subcaption{Pick}
      \end{minipage}
 \begin{minipage}[t]{0.29\linewidth}
        \includegraphics[width=\linewidth,trim=0cm 0cm 0cm 0cm,clip]{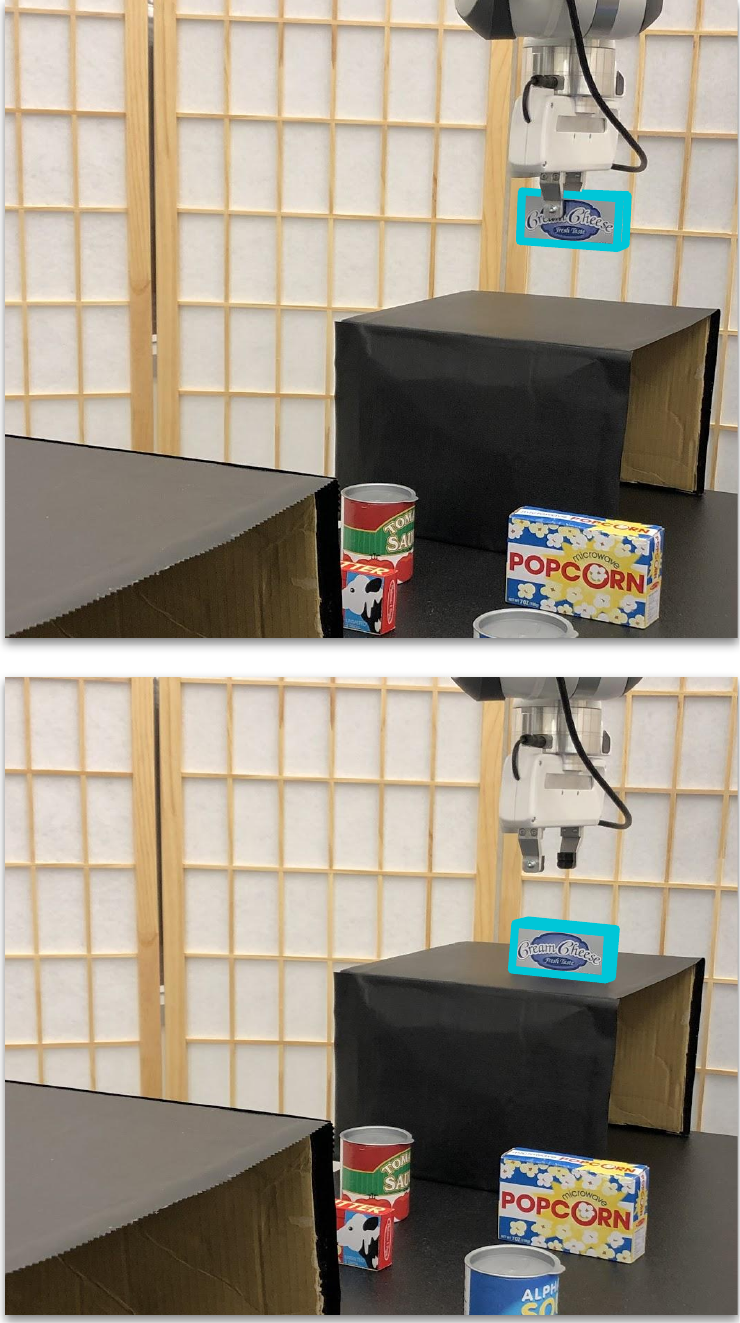}
        \subcaption{Place}
      \end{minipage}
     \begin{minipage}[t]{0.29\linewidth}
        \includegraphics[width=\linewidth,trim=0cm 0cm 0cm 0cm,clip]{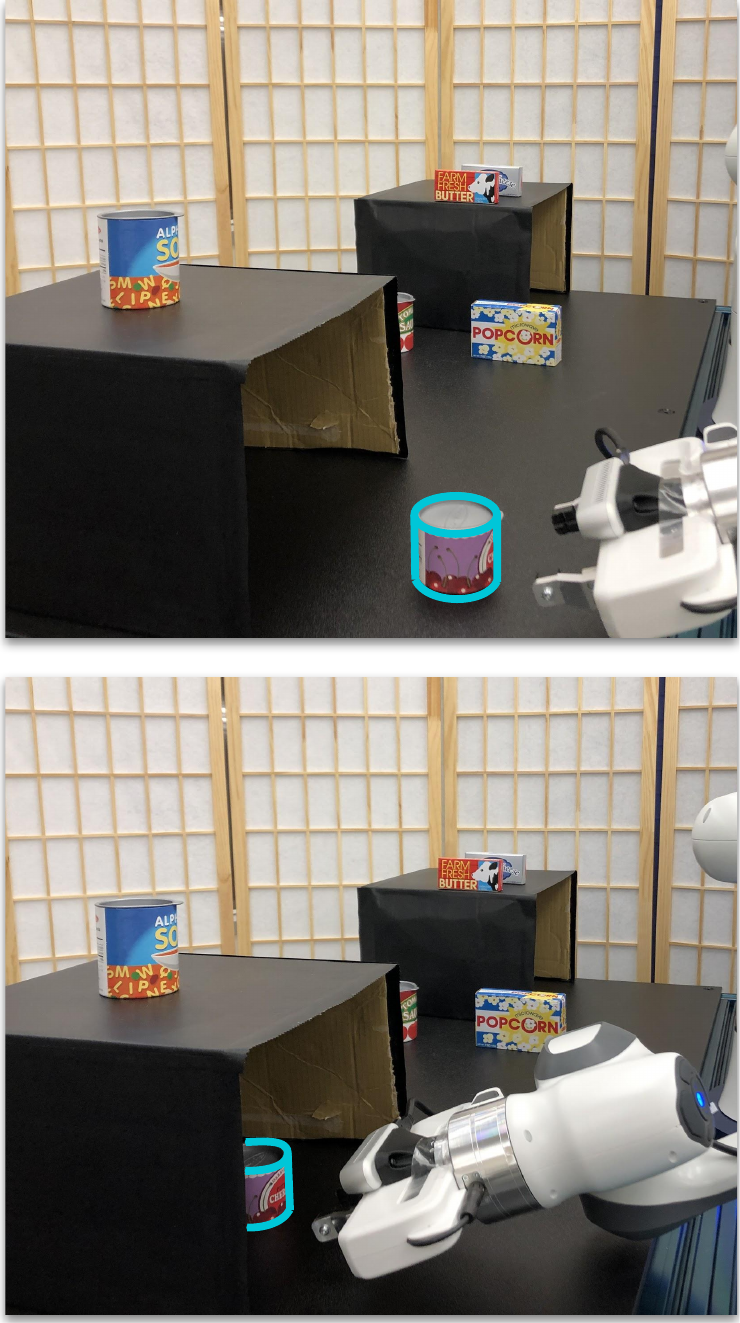}
      \subcaption{Push}        
      \end{minipage}   
    \vspace{-1mm}
    \caption{Examples of the motion primitives used in our storage tasks.}
    \label{fig:experiment}
    \vspace{-6mm}
\end{figure}

For experimental validation we designed a kitchen storage environment with two fixtures: shelf A on the left and shelf B on the right (see Figure~\ref{fig:overview}). 
In our tasks, the robot manipulates one or several objects based on the goal specification, where objects are specified to be inside the shelf, on the shelf, or on top of another object. We consider three types of motion primitives: \pick, \place, and \push, execution of which are shown in Fig~\ref{fig:experiment}. We define two regions on the table: open space in the middle and prepush regions in front of the shelves for objects to be pushed into a shelf. We use a selection of household objects from the HOPE dataset~\cite{tyree2019hope}, including three cans and three boxes, namely \texttt{Cherries}, \texttt{AlphabetSoup}, \texttt{TomatoSoup} and \texttt{Popcorn}, \texttt{CreamCheese}, \texttt{Butter}. We use DOPE~\cite{tremblay2018corl:dope, shi2020fast} to estimate poses of these objects for our real-world experiments.
We build up our simulated environments based on RoboVat~\cite{fang2019cavin}, a PyBullet-based simulation framework for robot manipulation, along with a photo-realistic rendering library NViSII~\cite{Morrical20nvisii}. 
The experiments are conducted with a 7-DoF Franka Emika Panda robot arm.



\paragraph{Logical Predicates} In our tasks, we have four logical predicates for describing the symbolic states of objects: \InGrasp~describes the state when the object is in the robot's gripper. \Clear~describes the state when the object is clear to be grasped, or clear to be stacked on. \In~for objects inside the shelf, and \On~for objects on a region, a shelf or another object. With predicates defined, we can directly provide goals in symbolic forms like \On (\texttt{Can}, \texttt{shelf A}).

\paragraph{Motion Primitive Parametrization} In our task, we parameterize \place~with $x$, $y$ locations in the environment, and discretize each dimension into 10 discrete bins. 
We parameterize \push~with $\theta$ for the angle to push and $d$ for the distance to travel along the pushing angle. We also discretize each dimension, and we have 21 discrete bins for each. 
For simplicity, we do not predict the orientation for picking an object with \pick~or the position for stacking an object with \place, but rather rely on the information from the geometric scene graph to determine the parameters. All motion trajectories are given by an RRT-planner.



\paragraph{Attributes for Grounded Scene Graphs} For \place, grounding $f_g$ assigns node features with its object type, object's current region, and its pose relative to target region from $\hat{g}$. For \push, grounded attributes for each node include its object type, its region, its pose relative to the object in $\hat{g}$, and its pose relative to shelf A.



\subsection{Data Collection and Training Details}

Our task planning modules are trained using a set of short demonstrations of the tasks. We only provide short demonstrations of how to place one object on a shelf, stack two objects on a shelf, or put one can into shelf A from initial configurations, resulting in 18 demonstration trajectories of tasks. For training our motion generation model, we collect all positive samples (successful execution) from sampled actions and create a dataset of 14,573 samples for \place~and 8,000 samples for \push~ in simulation.

We train both task planning and motion generation models with cross entropy loss. For all shared modules in \fpreimage~and \freach, we use a fully connected layer with Leaky ReLU activations. For our motion generation model, we have two-round message passing in GNN to obtain the embedding from grounded scene graph $\hat{s}_{g}$ and a fully-connected network to obtain embedding from subgoal $\hat{g}$. We use two-layer network for shared modules in the GNN model. Both embeddings are 32-dimensional vectors. After concatenating two embeddings, we have another two-layer network to predict parameters of motion primitives. Every layer in all modules has 256 hidden units.

\subsection{Evaluation Setup}
We evaluate our model on tasks with problem sizes ranging from two to six objects. In a task with $n$-object problem size, we have a goal specifying symbolic states of $n$ objects. We exclude stacking objects cases for two-object problem size as they are provided in the demonstration.

We have two major evaluation metrics: success rate and subgoal completion rate. Success rate is the percentage of successful completion of the whole task out of all trials. Subgoal completion rate is the number of subgoals completed over the number of total subgoals in goal $g$ to complete. Success rate indicates the overall performance of the model, and subgoal completion rate shows to what extent a task is completed even if it fails in the middle. We demonstrate the performance of our framework by thorough evaluation with ground-truth poses in simulation, and evaluate its real-world applicability on a real robot with visual observation included.

\subsection{Simulation Experiments with Ground-Truth Poses}

We randomly generate 100 symbolic goal specifications for each problem size, and we evaluate five times repeatedly for each goal specification. To examine the strengths of our regression planning as opposed to direct forward prediction, we compare with a baseline method which directly predicts the next subgoal conditioned on $g$ and $s_g$ using GNN models. 

The performance of our approach is reported in Table~\ref{tab:gt-results}, along with two motion generation baselines: single-shot random sampling approach (\texttt{Random}), which uniformly samples a parameter from the parameter space, and a multi-layer perceptron approach (\texttt{MLP}), where we flatten all node attributes and edge attributes into a single vector, and use a fully-connected network of the same capacity as the GNN. We omit the standard deviations as all the standard deviation values are below $0.04$. Our results indicate that direct subgoal prediction has \textbf{zero} success rate for all problem sizes (thus not presented in the table), due to its inability of generalization to tasks longer than the training demonstrations. 

As the result suggests, our graph-based motion generation model is effective for single-shot prediction of $\theta$ compared to the random sampling baseline, and the relational inductive biases in GNNs boost performance as compared to the MLP baseline. Failure modes in our approach are primarily due to execution failure, and the rest are overstep failure. Execution failure refers to failure due to motion execution, and overstep failure is failure because of executing the same primitives in an infinite loop or never predicting a reachable subgoal within maximal steps of $50$. We have $7$\% execution failure for tasks with 2 objects, all the way to 33\% for tasks with 6 objects. 
Overstep failure never exceeds 1\%.

A major advantage of our approach over classical TAMP method is its efficient inference for long-horizon tasks. The use of learning models substantially reduces the amount of search problems it has to solve.
Table~\ref{tab:time} shows a comparison of computation time against the state-of-the-art TAMP solver PDDLStream~\cite{garrett2020pddlstream}. We found that our method significantly accelerates the inference procedure on both task and motion levels, especially for task planning, where the neuro-symbolic planner is four orders of magnitude faster than the search-based logic solver with PDDL.

\begin{table}
  \centering
    \caption{\label{tab:gt-results} Success rate / subgoal completion rate in simulated environments with ground-truth poses: random sampling (Random), fully-connected network (MLP), and our method (Ours).}
   \resizebox{0.85\linewidth}{!}{  
  \begin{tabular}{llll}
    \toprule
    \textbf{Problem Size} & \textbf{Random} & \textbf{MLP} & \textbf{Ours~(GNNs)}\\
    \midrule
    2 objects & $0.71$ / {$0.84$} & {$0.85$} / {$0.91$} & {$\mathbf{0.92}$} / $0.95$ \\
    3 objects & $0.60$ / $0.81$ & {$0.75$} / $0.87$ & {$\mathbf{0.89}$} / $0.95$\\
    4 objects & $0.50$ / $0.78$ & {$0.69$} / $0.84$ & {$\mathbf{0.84}$} / $0.92$ \\
    5 objects & $0.35$ / $0.67$ & {$0.56$} / $0.76$ & {$\mathbf{0.73}$} / $0.87$\\
    6 objects & $0.18$ / $0.56$ & $0.44$ / $0.67$ & {$\mathbf{0.67}$ / $0.84$} \\
    \bottomrule
  \end{tabular}
  }
  \vspace{-2mm}
  \end{table}

\begin{table}[t]
\centering
\caption{Computation time (in seconds) of the PDDLStream solver versus Ours (including RRT motion planning). Our model runs on Intel Xenon 2.20 GHz with RTX 2080.}.
\label{tab:time}
  \resizebox{0.80\linewidth}{!}{  
  \begin{tabular}{lcccc}
    \toprule
    \textbf{Problem Size} & \multicolumn{2}{c}{\textbf{PDDLStream}~\cite{garrett2020pddlstream}} & \multicolumn{2}{c}{\textbf{Ours}} \\
    & Task & Motion & Task & Motion\\
    \midrule
    2 objects & $127.9$ & $23.5$ & $0.15$ & $18.1$ \\
    3 objects & $251.7$ & $35.7$ & $0.13$ & $27.2$ \\
    4 objects & $251.2$ & $62.9$ & $0.22$ & $36.5$ \\
    5 objects & $322.2$ & $107.0$ & $0.29$ & $42.9$ \\
    6 objects & $424.2$ & $276.9$ & $0.32$ & $49.4$ \\
    \bottomrule
  \end{tabular}
 }
  \vspace{-5mm}
\end{table}

\subsection{Real Robot Experiments with Visual Observations}
We evaluate our approach on the real robot with raw visual input. To achieve robust pose estimation results for computing the geometric scene, we take multiple images of different viewpoints to cover the workspace. We evaluate 34 trials in total (10, 10, 8, 3, 3 trials for 2--6 objects, respectively).  The success rate / subgoal completion rate for each problem size is 0.80/0.90, 0.80/0.90, 0.63/0.88, 0.67/0.93, 0.33/0.89.  The overall success rate is $\mathbf{70.6}\%$ and the subgoal completion rate is $\mathbf{89.6}\%$. The gap between the success rate and subgoal completion rate is due to execution failure of the motion primitives for the last few subgoals in larger problem sizes. We show qualitative examples of our real-world experiments in the supplementary video.

%% file: 6-conclusion.tex
In this paper we present a visual hierarchical planning algorithm for long-horizon manipulation tasks. Our framework integrates neuro-symbolic task planning and graph-based motion generation on graph-based scene representations. The essence of our method lies in the two-level abstractions of a manipulation scene with \emph{geometric scene graphs} and \emph{symbolic scene graphs}. We show the good generalization ability to longer and novel tasks with models trained on a set of short demonstration. We quantitatively evaluate our approach in simulation and real-world experiments. Two promising directions are left for future work. One is how to learn the symbol mapping function in a self-supervised way from demonstration, which can further automate the framework and learning process. The other direction is to develop a more flexible form of scene graph representations that can represent articulated and deformable objects.